\documentclass{article}

\usepackage{arxiv}

\usepackage[utf8]{inputenc}
\usepackage[T1]{fontenc}
\usepackage{hyperref}
\usepackage{url}
\usepackage{booktabs}
\usepackage{amsfonts}
\usepackage{nicefrac}
\usepackage{microtype}
\usepackage{graphicx}

\usepackage{amsmath}
\usepackage{xcolor}
\usepackage{wrapfig}
\usepackage{placeins} 
\usepackage{float}    

\graphicspath{{./}{../}}

\hypersetup{colorlinks=true, linkcolor=black, citecolor=black, urlcolor=black}

\title{DC-Mamba: BI-TEMPORAL DEFORMABLE ALIGNMENT AND SCALE-SPARSE ENHANCEMENT FOR REMOTE SENSING CHANGE DETECTION}

\author{
  Min Sun \\
  School of Materials Science and Engineering \\
  Hainan University \\
  \And
  Fenghui Guo\thanks{Corresponding author.} \\
  School of Materials Science and Engineering \\
  Hainan University \\
}

\begin{document}
\maketitle

\begin{abstract}
Remote sensing change detection (RSCD) is vital for identifying land-cover changes,
yet existing methods, including state-of-the-art State Space Models (SSMs), often
lack explicit mechanisms to handle geometric misalignments and struggle to
distinguish subtle, true changes from noise. To address this, we introduce DC-Mamba,
an "align-then-enhance" framework built upon the ChangeMamba backbone. It integrates
two lightweight, plug-and-play modules: (1) Bi-Temporal Deformable Alignment (BTDA),
which explicitly introduces geometric awareness to correct spatial misalignments at
the semantic feature level; and (2) a Scale-Sparse Change Amplifier (SSCA), which
uses multi-source cues to selectively amplify high-confidence change signals while
suppressing noise before the final classification. This synergistic design first
establishes geometric consistency with BTDA to reduce pseudo-changes, then leverages
SSCA to sharpen boundaries and enhance the visibility of small or subtle targets.
Experiments show our method significantly improves performance over the strong
ChangeMamba baseline, increasing the F1-score from 0.5730 to 0.5903 and IoU from
0.4015 to 0.4187. The results confirm the effectiveness of our "align-then-enhance"
strategy, offering a robust and easily deployable solution that transparently
addresses both geometric and feature-level challenges in RSCD.
\end{abstract}



\section{Introduction}
\label{sec:intro}
\begin{figure}[H]
\centering
\includegraphics[width=8.5cm]{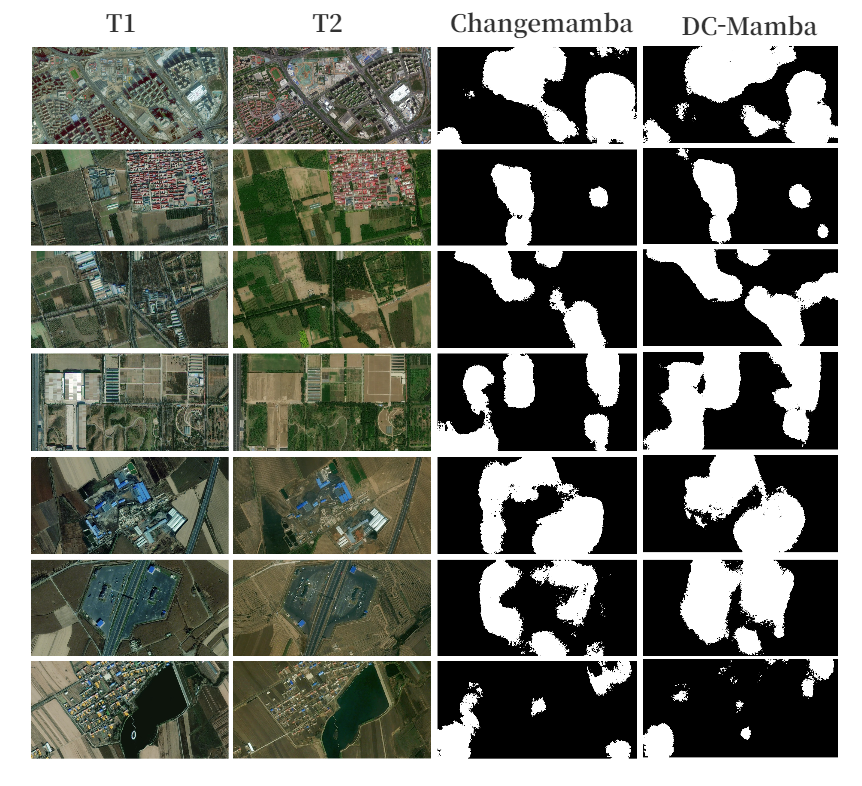}
\caption{Qualitative comparisons across diverse scenes: DC-Mamba shows crisper boundaries, better small-object recovery, and fewer pseudo-changes than ChangeMamba (prior SOTA)\cite{10565926}; F1/IoU improve from 0.5730/0.4015 to 0.5903/0.4187.}
\label{fig:mainall}
\end{figure}

Remote Sensing Change Detection (RSCD) aims to identify and delineate land cover
changes from bi-temporal images, thereby supporting disaster assessment, urban
planning, and environmental and agricultural monitoring. Despite rapid advancements
in deep learning-driven techniques, achieving stable and fine-grained pixel-level
differentiation remains challenging in large-scale, heterogeneous scenes due to
inter-temporal misalignment, radiometric and appearance variations, class-imbalanced
scale changes, and the weak visibility of boundaries and textures. These challenges
have motivated boundary-aware and lightweight designs in recent literature \cite{10633742,tinycd2022nca}. Qualitative comparisons are shown in Fig.~\ref{fig:mainall}.

Current mainstream deep learning methods for change detection originate from the
fully convolutional siamese network \cite{8451652}, which established the basic paradigm of
fusing multi-scale features through an encoder-decoder architecture. However, this
direct feature fusion was quickly found to be coarse, leading to semantic gaps and
blurred boundaries. This spurred a series of methods attempting to optimize the
fusion process within the CNN framework, such as through dense connections \cite{9355573},
spatio-temporal attention \cite{rs12101662}, deep metric supervision \cite{9555146}, and explicit feature
interaction \cite{fang2022changer}. Although these efforts
improved performance, they are fundamentally limited by the local receptive field of
convolutions. To break this locality barrier, another line of research has adopted
Transformers \cite{9491802,9883686,9736956,9761892} to effectively model long-range spatio-temporal dependencies.
In parallel, unsupervised and video-style formulations have also been explored \cite{8824216,9975266}. However, as recent research \cite{10565926} has pointed out, this global modeling capability
comes at the cost of quadratic computational complexity and the potential loss of
fine details due to image tokenization. This leaves a critical challenge: how to
achieve global contextual awareness and high computational efficiency simultaneously
without sacrificing local feature precision. Recently, State Space Models (SSMs),
represented by the state-of-the-art ChangeMamba \cite{10565926}, have emerged as a third
promising direction, striking a compelling balance between global context modeling
and linear complexity. However, while it establishes a new performance benchmark, its
inherent sequential modeling nature does not explicitly address the prevalent issues
of geometric misalignment and the separability of subtle changes in complex remote
sensing scenes; single-temporal supervised CD has also been investigated \cite{changestar2021}, and
foreground-aware segmentation frameworks provide complementary insights for boundary quality \cite{zheng2023farseg++}.

\begin{enumerate}
  \item Introducing DC-Mamba, which upgrades the existing ChangeMamba model through a novel "align-then-enhance" framework, significantly
        improving performance in complex scenarios while maintaining high efficiency.
  \item Designing BTDA, the core component of DC-Mamba, which performs robust and
        interpretable deformable alignment without extra supervision to mitigate geometric
        inconsistencies.
  \item Designing SSCA, the second core component of DC-Mamba, which is a lightweight,
        plug-and-play amplifier that selectively enhances true change signals and sharpens
        target boundaries.
\end{enumerate}
\section{PROPOSED METHOD}
\label{sec:format}

\subsection{Overall Framework}
\label{ssec:overall}

The overall framework we propose is named DC-Mamba, which is built upon the advanced siamese ChangeMamba backbone and follows a core design philosophy of "align-then-enhance." This framework achieves systematic enhancement by integrating two lightweight, plug-and-play modules we designed: BTDA (Bi-temporal Deformable Alignment) and SSCA (Scale-Sparse Change Amplifier).

Specifically, the BTDA module first performs explicit alignment of cross-temporal features at a high semantic level to correct for geometric misalignments caused by factors such as imaging angles and temporal differences. Subsequently, the aligned features are fed into the subsequent network for change modeling. Before entering the final classification head, the SSCA module selectively amplifies and enhances high-confidence change signals to sharpen boundaries and highlight subtle changes.

The entire workflow keeps the backbone network and task head unchanged, ensuring low coupling and low overhead. This design greatly facilitates tiled inference and cross-regional deployment of the model on large-scale remote sensing data.

\subsection{BTDA: Bi-Temporal Deformable Alignment}
\label{ssec:btda}

Although ChangeMamba achieves excellent global context modeling through its sequential scanning mechanism, the inherent nature of this "one-dimensional scan" means it naturally lacks the ability to explicitly perceive two-dimensional spatial geometric relationships. Consequently, when slight rotations, scaling, or non-rigid deformations exist between bi-temporal images due to factors like camera angles or terrain, its 1D scanning mechanism struggles to effectively model this 2D offset, easily producing a large number of pseudo-changes at the edges of unaligned objects.

\begin{figure}[htb]
\begin{minipage}[b]{1.0\linewidth}
  \centering
  \centerline{\includegraphics[width=8.5cm]{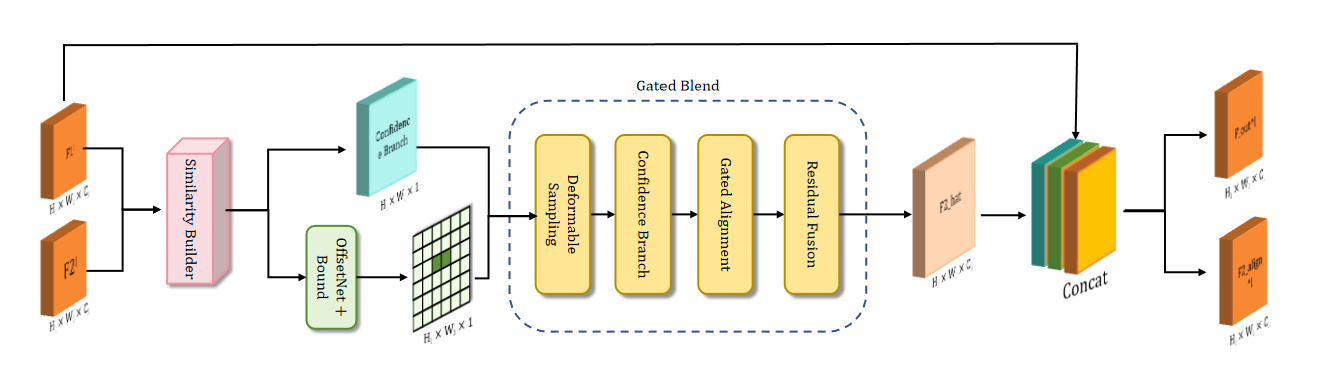}}
\end{minipage}
\caption{BTDA: bounded offsets with a scalar gate align bi-temporal high-level features to reduce pseudo-changes.}
\label{fig:btda}
\end{figure}

To address this core shortcoming, we designed the BTDA (Bi-temporal Deformable Alignment) module as the first step in our "align-then-enhance" framework. Its goal is to explicitly introduce the capability to handle geometric inconsistencies for ChangeMamba within the high-level semantic feature space. Specifically, drawing inspiration from deformable convolution\cite{zhu2019dcnv2}, we use a lightweight convolutional head on the concatenated bi-temporal features to dynamically predict a 2D offset field $\Delta^{(l)}$ and a scalar gate $\lambda^{(l)}$. An overview is shown in Fig.~\ref{fig:btda}.

Crucially, to avoid disrupting the powerful semantic features already learned by ChangeMamba, our alignment is restrained and precise: through a gating mechanism, the network can adaptively drive the gate value $\lambda^{(l)}$ towards 0 in well-aligned regions, thereby weakening or even skipping unnecessary deformations to improve flexibility and stability; furthermore, we impose an amplitude limit on the predicted offset field (see Eq.~\eqref{eq:offset_bound}) to prevent extreme distortions that could destroy the feature topology; finally, we introduce an $\ell_2$ regularization term $\mathcal{L}_{\text{off}}$ (see Eq.~\eqref{eq:l_off}) to encourage smaller deformations, ensuring a smoother and more physically plausible alignment process.

\begin{equation}
\widetilde{\Delta}^{(l)} = \delta_{\max}^{(l)} \cdot \tanh\!\big(\Delta^{(l)}\big)
\label{eq:offset_bound}
\end{equation}

\begin{equation}
\mathcal{L}_{\text{off}} = \sum_{l} \big\|\widetilde{\Delta}^{(l)}\big\|_2
\label{eq:l_off}
\end{equation}

Finally, as shown in Eq.~\eqref{eq:deform_resample}, we use the learned gated offset field to resample the post-change features via Deformable Sampling\cite{zhu2019dcnv2}, obtaining new features that are aligned with the pre-change features. This process is like installing a plug-and-play "geometric corrector" for ChangeMamba, allowing it to eliminate major geometric noise before conducting change analysis, thus laying a solid foundation for subsequent precise change identification.

\begin{equation}
\widehat{F}_{t\prime}^{(l)}(x) = \text{DeformSample}\!\Big( F_{t\prime}^{(l)},\, x + \lambda^{(l)}\,\widetilde{\Delta}^{(l)}(x) \Big)
\label{eq:deform_resample}
\end{equation}

\subsection{SSCA: An Intelligent Signal Amplifier for Fine-Grained Changes}
\label{ssec:ssca}

Although BTDA corrects for macroscopic geometric inconsistencies and provides a clean foundation for change analysis, another core challenge remains: how to accurately distinguish real, often subtle, land-cover changes from meaningless feature fluctuations (e.g., due to lighting or seasonal differences) in the aligned features? Simply calculating feature differences is sensitive to noise and can easily lead to “holes” within change regions or produce blurry, discontinuous responses at the edges.

\begin{figure}[!t]
\begin{minipage}[b]{1.0\linewidth}
  \centering
  \centerline{\includegraphics[width=8.5cm]{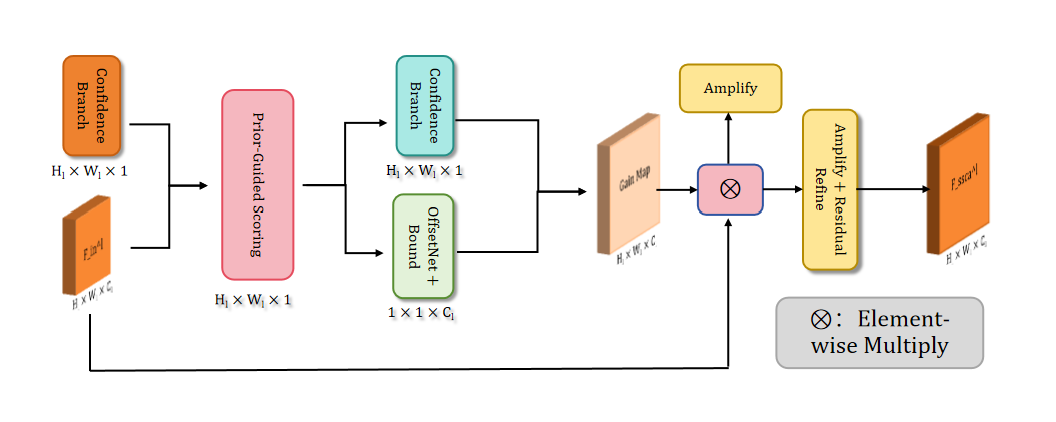}}
\end{minipage}
\caption{SSCA: a spatial gate and channel weights drive residual amplification to emphasize reliable changes and sharpen boundaries.}
\label{fig:ssca}
\end{figure}

To tackle this challenge, we designed the second step of the “align-then-enhance” framework: SSCA (Scale-Sparse Change Amplifier). Its core mission is to act as a precision change signal amplifier: before the final classification, it proactively identifies and enhances high-confidence true change features while suppressing noise, thereby sharpening the boundaries of change targets and bringing “focus” to small or subtle changes that are easily overlooked. The overall pipeline is illustrated in Fig.~\ref{fig:ssca}.

SSCA’s precision amplification mechanism works in two steps: first, it determines “where” to amplify, and then it decides “what” to amplify.

1) Locating Change Regions (Spatial Gating): To accurately locate potential change regions, we do not rely on a single feature difference. Instead, we aggregate four complementary Multi-source Change Cues, as shown in Eq.~\eqref{eq:cue_stack}. These cues include: A) Absolute Feature Difference $|F_t - F_{t'}|$ as the most direct evidence of change; B) Bi-temporal Spatial Gradients $\nabla F_t,\, \nabla F_{t'}$ to capture texture/structure changes with edge sensitivity; C) Differentiable Structural Similarity (DSSIM) $\text{DSS}(F_t, F_{t'})$, measuring structural changes at a perceptual level and being robust to illumination variations. We stack these cues and feed them into a lightweight convolutional network to generate a spatial gate map $M$. Each pixel value in $M$ represents the likelihood of a true change at that location, thus guiding the model on “where” to apply stronger attention.

\begin{equation}
\mathcal{C} = \big\{\, \lvert F_t - \widehat{F}_{t\prime} \rvert,\ \nabla F_t,\ \nabla 
\widehat{F}_{t\prime},\ \text{DSS}(F_t,\widehat{F}_{t\prime}) \,\big\}
\label{eq:cue_stack}
\end{equation}

2) Enhancing Change Features (Channel Recalibration): After identifying the locations of change, the model also needs to know which feature channels contribute most to the concept of “change.” To this end, we introduce a channel attention mechanism, as shown in Eq.~\eqref{eq:channel_weights}. By applying global average pooling (GAP) and two fully connected layers ($W_1, W_2$) to the fused features $F$, we compute a weight $s$ for each channel. This weight vector $s$ reflects the importance of different feature channels for identifying change, guiding the model on “what” to amplify.

\begin{equation}
s = \sigma\!\Big( W_2\, \mathrm{ReLU}\!\big( W_1\, \mathrm{GAP}(F) \big) \Big), \quad s \in [0,1]^C
\label{eq:channel_weights}
\end{equation}

Finally, as shown in Eq.~\eqref{eq:residual_amplify}, SSCA combines the spatial gate $M$ and the channel weights $s$ to perform a Residual Amplification on the original features $F$. Here, $\alpha$ is a learnable scalar that controls the overall intensity of the amplification. This design makes the enhancement process targeted and adaptive: only signals that are considered to be in a change region spatially and are deemed important features channel-wise will be significantly enhanced.

\begin{equation}
F' = F + \alpha \cdot (M \odot s) \odot F
\label{eq:residual_amplify}
\end{equation}




\section{Experimental Results and Analysis}
\label{sec:pagestyle}

\subsection{Experimental Setup}
\label{ssec:exp_setup}

Environment and implementation: All experiments are conducted on a Tesla V100-SXM2-16GB GPU with CUDA~12.6 and Python~3.10. All baselines use their official implementations and publicly released pretrained weights under the same data split and input resolution; each model is trained for 1800 iterations. Input images are resized to 1280$\times$630.

Data and split: The dataset contains 400 images in total (including both temporal phases \(T1\) and \(T2\)). We use a 9:1 train:val split, and an independent test set of 50 images. Unless otherwise specified, we report six metrics: Precision (P), Recall (R), Overall Accuracy (OA), F1, Intersection over Union (IoU), and Cohen's Kappa (Kappa).

\subsection{Overall Results}
\label{ssec:overall_results}

Table 1 extends the comparison by adding our DC-Mamba (BTDA+SSCA). DC-Mamba achieves the best \( \text{F1} \) (0.5903), \( \text{IoU} \) (0.4187), and \( \kappa \) (0.4507) among all methods, and the highest Precision (0.6871), while maintaining competitive Recall (0.5174). Compared with the strong ChangeMamba baseline, DC-Mamba improves \( \text{F1} \) by +0.0173 and \( \text{IoU} \) by +0.0172 with a modest decrease in \( \text{OA} \) (-0.0173) and Recall (-0.0416). These results substantiate the align-then-enhance design: BTDA improves geometric consistency and suppresses registration-induced pseudo changes, and SSCA concentrates model capacity on reliable spatial-channel cues to sharpen boundaries and better recover small objects. In class-imbalanced RSCD, \( \text{IoU} \) and \( \kappa \) are more indicative of segmentation quality than \( \text{OA} \); DC-Mamba leads on both, producing cleaner and more consistent change maps.

\begin{table}[!htbp]
\centering
\caption{Quantitative comparison with representative baselines, including our DC-Mamba. Higher is better.}
\label{tab:baseline_cmp}
\scriptsize
\setlength{\tabcolsep}{5.2pt}
\renewcommand{\arraystretch}{1.1}
\resizebox{0.9\columnwidth}{!}{
\begin{tabular}{lcccccc}
\hline
Method & P & R & OA & F1 & IoU & Kappa \\
\hline
ChangeMamba\cite{10565926} & 0.5817 & 0.5590 & 0.8049 & 0.5730 & 0.4015 & 0.4467 \\
ChangeFormerV6\cite{bandara2022changeformer} & 0.4676 & 0.5393 & 0.7920 & 0.5009 & 0.3342 & 0.3426 \\
Changer\cite{fang2022changer} & 0.6539 & 0.4392 & 0.7370 & 0.5255 & 0.3563 & 0.3531 \\
ChangeStar\cite{changestar2021} & 0.2950 & 0.8780 & 0.5055 & 0.4416 & 0.2834 & 0.1624 \\
BIT\cite{9491802} & 0.4111 & 0.3995 & 0.7312 & 0.4052 & 0.2541 & 0.2317 \\
TinyCD\_v2\cite{tinycd2022nca} & 0.4599 & 0.4650 & 0.7619 & 0.4624 & 0.3008 & 0.3095 \\
STANet\cite{rs12101662} & 0.5084 & 0.4655 & 0.7605 & 0.4860 & 0.3210 & 0.3303 \\
SNUNet\cite{9355573} & 0.3689 & 0.4099 & 0.7412 & 0.3883 & 0.2410 & 0.2248 \\
DC-Mamba (Ours) & \textcolor{red}{0.6871} & 0.5174 & 0.7876 & \textcolor{red}{0.5903} & \textcolor{red}{0.4187} & \textcolor{red}{0.4507} \\
\hline
\end{tabular}}
\end{table}
\FloatBarrier

\subsection{Ablation Studies and Analysis}
\label{ssec:ablation}

As shown in Table 2, introducing BTDA at high semantic levels improves geometric consistency and reduces pseudo changes, which primarily translates into a precision gain (\( \text{P}: 0.5817 \rightarrow 0.6218 \)) with nearly unchanged recall (\( \text{R}: 0.5590 \rightarrow 0.5549 \)), yielding steady gains in \( \text{F1} \), \( \text{IoU} \), and \( \kappa \). This confirms that mitigating cross-temporal misalignment at high-level features is an effective way to suppress registration-induced noise without sacrificing coverage.

\begin{table}[!htbp]
\centering
\scriptsize
\setlength{\tabcolsep}{5.2pt}
\renewcommand{\arraystretch}{1.1}
\caption{Ablation on the proposed modules. Baseline is ChangeMamba.}
\label{tab:ablation}
\resizebox{0.9\columnwidth}{!}{
\begin{tabular}{lcccccc}
\hline
Variant & P & R & OA & F1 & IoU & Kappa \\
\hline
DC-Mamba & 0.6871 & 0.5174 & 0.7876 & 0.5903 & 0.4187 & 0.4507 \\
-SSCA & 0.6218 & 0.5549 & 0.8047 & 0.5864 & 0.4149 & 0.4592 \\
-BTDA-SSCA & 0.5817 & 0.5590 & 0.8049 & 0.5730 & 0.4015 & 0.4467 \\
\hline
\end{tabular}}
\end{table}

Stacking SSCA before the classifier further boosts precision (\( \text{P}: 0.6218 \rightarrow 0.6871 \)) while reducing recall (\( \text{R}: 0.5549 \rightarrow 0.5174 \)), leading to modest improvements in \( \text{F1} \) and \( \text{IoU} \). The selective amplification and sparsity prior favor high-confidence regions and sharper boundaries, which can suppress low-confidence noise but also filter out subtle, low-contrast changes. If recall is prioritized in a particular application, this trade-off can be adjusted by relaxing the gate sparsity, lowering the classification threshold, or strengthening boundary-consistency terms to recover faint changes.

Overall, the align-then-enhance design is validated: BTDA first cleans up geometry and reduces pseudo-change floor, and SSCA subsequently concentrates model capacity on informative channels and regions. Compared to using either module alone, the serial composition achieves a more transparent and stable improvement trajectory with minimal engineering overhead.

\subsection{Qualitative Observations and Efficiency}
\label{ssec:qual_eff}

Qualitative inspections (omitted here for brevity) show better cross-temporal contour alignment of unchanged objects with BTDA, and sharper boundaries plus improved small-object recovery with SSCA. Both modules are lightweight and plug-and-play, adding minor latency and memory overhead relative to the backbone, which supports tiled large-area inference in resource-constrained deployments.

\section{Conclusion}
\label{sec:typestyle}

We present DC-Mamba, an align-then-enhance pipeline for remote sensing change detection. BTDA performs high-level bi-temporal deformable alignment to suppress registration-induced pseudo changes, while SSCA uses spatial gating and channel reweighting to residually amplify reliable change cues, sharpening boundaries and improving small-object visibility. Under the same protocol, DC-Mamba improves F1 from 0.5730 to 0.5903 and IoU from 0.4015 to 0.4187, and attains the highest Precision (0.6871) and Kappa (0.4507) among compared methods. Both modules are lightweight and plug-and-play; ablations verify their complementary and interpretable effects. The precision/recall trade-off introduced by SSCA can be tuned via gate sparsity or decision thresholds to meet application needs.

\vfill\pagebreak
\label{sec:refs}
\bibliographystyle{IEEEbib}
\nocite{*}
\bibliography{refs}

\end{document}